\documentclass[11pt]{article}
\usepackage[left=2.2cm,right=2.2cm,top=2.5cm,bottom=2.5cm,headheight=40pt,headsep=20pt, a4paper]{geometry}
\usepackage[utf8]{inputenc}
\usepackage[T1]{fontenc}

\usepackage[round, comma, sort&compress]{natbib}
\usepackage{xspace}
\usepackage{array}
\usepackage{courier}
\usepackage{xcolor}
\usepackage{multirow}
\usepackage{makecell}
\usepackage{caption}
\usepackage{subcaption}
\usepackage{longtable}
\usepackage{booktabs}
\usepackage{array}
\usepackage{graphicx}
\usepackage{verbatim}
\usepackage{authblk}
\usepackage{newfloat}

\definecolor{darkgreen}{RGB}{0,100,0}
\usepackage[hidelinks=false,
            hyperindex,
            breaklinks,
            colorlinks=true,
            urlcolor=darkgreen,
            linkcolor=blue,
            citecolor=blue,
            backref=page,
            pdfauthor={Faruk Ahmed},
            pdftitle={PolyPath -- Adapting a Large Multimodal Model for Multi-slide Pathology Report Generation},
            pdfsubject={PolyPath},
            pdfproducer={Latex with hyperref},
            pdfborder={0 0 0}]{hyperref}

\newcommand{\tthnoisy}{\textsc{DS1-Noisy}\xspace}

\newcommand{\pathalign}{\textsc{PathAlign}\xspace}
\newcommand{\ourmodel}{\textsc{PolyPath}\xspace}
\newcommand{\randommodel}{\textsc{SS-Random}\xspace}
\newcommand{\llmmodel}{\textsc{SS-LLM}\xspace}

\title{\textbf{\ourmodel: Adapting a Large Multimodal Model for Multi-slide Pathology Report Generation}}

\author{Faruk Ahmed$^\dagger$}
\author{Lin Yang}
\author{Tiam Jaroensri}
\author{Andrew Sellergren}
\author{Yossi Matias}
\author{\authorcr Avinatan Hassidim}
\author{Greg S. Corrado}
\author{Dale R. Webster}
\author{Shravya Shetty}
\author{\authorcr Shruthi Prabhakara}
\author{Yun Liu}
\author{Daniel Golden$^{\dagger *}$}
\author{Ellery Wulczyn$^{\dagger *}$}
\author{David F. Steiner$^{\dagger *}$}

\affil{\small Google Research}

\date{}

\begin{document}
\maketitle

\renewcommand*{\thefootnote}{\fnsymbol{footnote}}
\footnotetext[1]{These authors co-supervised the work.}
\footnotetext[2]{Contact: \{farukahmed, dangolden, ewulczyn, davesteiner\} AT google.com.}
\raggedbottom

\begin{abstract}
The interpretation of histopathology cases underlies many important diagnostic and treatment decisions in medicine. Notably, this process typically requires pathologists to integrate and summarize findings across multiple slides per case. Existing vision-language capabilities in computational pathology have so far been largely limited to small regions of interest, larger regions at low magnification, or single whole-slide images (WSIs). This limits interpretation of findings that span multiple high-magnification regions across multiple WSIs. By making use of Gemini 1.5 Flash, a large multimodal model (LMM) with a 1-million token context window, we demonstrate the ability to generate bottom-line diagnoses from up to 40,000 $768 \times 768$ pixel image patches from multiple WSIs at 10X magnification. This is the equivalent of up to 11 hours of video at 1 fps. Expert pathologist evaluations demonstrate that the generated report text is clinically accurate and equivalent to or preferred over the original reporting for 68\% (95\% CI: [60\%, 76\%]) of multi-slide examples with up to 5 slides. While performance decreased for examples with 6 or more slides, this study demonstrates the promise of leveraging the long-context capabilities of modern LMMs for the uniquely challenging task of medical report generation where each case can contain thousands of image patches.
\end{abstract}

\section{Introduction}
\label{sec:introduction}
Recent applications of vision--language modeling in digital histopathology have been predominantly designed to generate text describing individual regions of interest extracted from a single digitized histopathology image, or Whole Slide Image (WSI). An emerging line of research approaches the more practical clinical use case of slide-level text generation~\citep{chen2024wsicaption,ahmed2024pathalign}. However, in the typical clinical use case, there can be multiple biological tissue parts associated with a case, with each part having multiple slides. Pathologists write up a report summarizing their part-level diagnostic findings by microscopically reviewing each of the available slides per part and integrating information across these slides. This many-to-one relationship of slides to clinical descriptions is a recognized challenge for vision--language modeling in this space~\citep{ahmed2024pathalign}. The common approach taken in recent literature is to restrict modeling and analysis to single-slide cases or to manually identify a single slide within a case or part that is most representative of the clinical findings in reports~\citep{zhou2024pathm3,guo2024histgen,chen2024wsicaption,shaikovski2024prism,xu2024whole,ahmed2024pathalign}. This strategy of selecting representative slides was also adopted in constructing one of the most widely used histopathology datasets, TCGA~\citep{cooper2018pancancer}. \\ \\
While the natural solution for multi-slide report generation is to perform vision--language modeling using all available slides as input, this approach remains underexplored in existing literature, likely due to computational challenges involved in processing multiple WSIs. Two recent works are of note. The HistoGPT vision--language model~\citep{tran2024generating} is capable of handling multiple slides per patient at 10X magnification (corresponding to approximately 1 micron per pixel). The work focused exclusively on dermatopathology cases, and the model was evaluated within a human-in-the-loop “expert-guidance” mode, where a human expert prompts the model with their preliminary impression. With this preliminary impression, the model was shown to generate reports on par with those written by human experts. In another line of work,~\citet{tan2024clinical} proposed a multimodal model to generate reports using multiple slides with a total of 7422 slides from 1991 patients, covering kidney and colon cases. Slides at lower magnifications of 1.25X and 5X were used (corresponding to resolutions of approximately 8 and 2 microns per pixel respectively), which can be limiting since pathologists use higher magnifications for many diagnostic tasks~\citep{kim2020re}. While the work did not conduct evaluation with human experts, the reports are pre-processed into a structured format for typical kidney and colon diagnoses and evaluated using natural language generation (NLG) metrics. \\ \\
Recent developments in long-context multimodal modeling~\citep{team2024gemini} enables us to explore the computationally challenging task of report generation from multiple WSIs. In this work, we process up to 50 slides per part at a 10X magnification using the dataset in~\citet{ahmed2024pathalign}. The data consists of approximately 350K WSIs, spanning a wider range of tissue types and conditions than prior multi-slide works~\citep{tran2024generating,tan2024clinical}. We evaluate the part-level report generation performance of our model with both NLG metrics and expert evaluation.

\begin{figure}[t]
    \centering
    \includegraphics[trim={0 5cm 0cm 1.5cm},clip,scale=0.5]{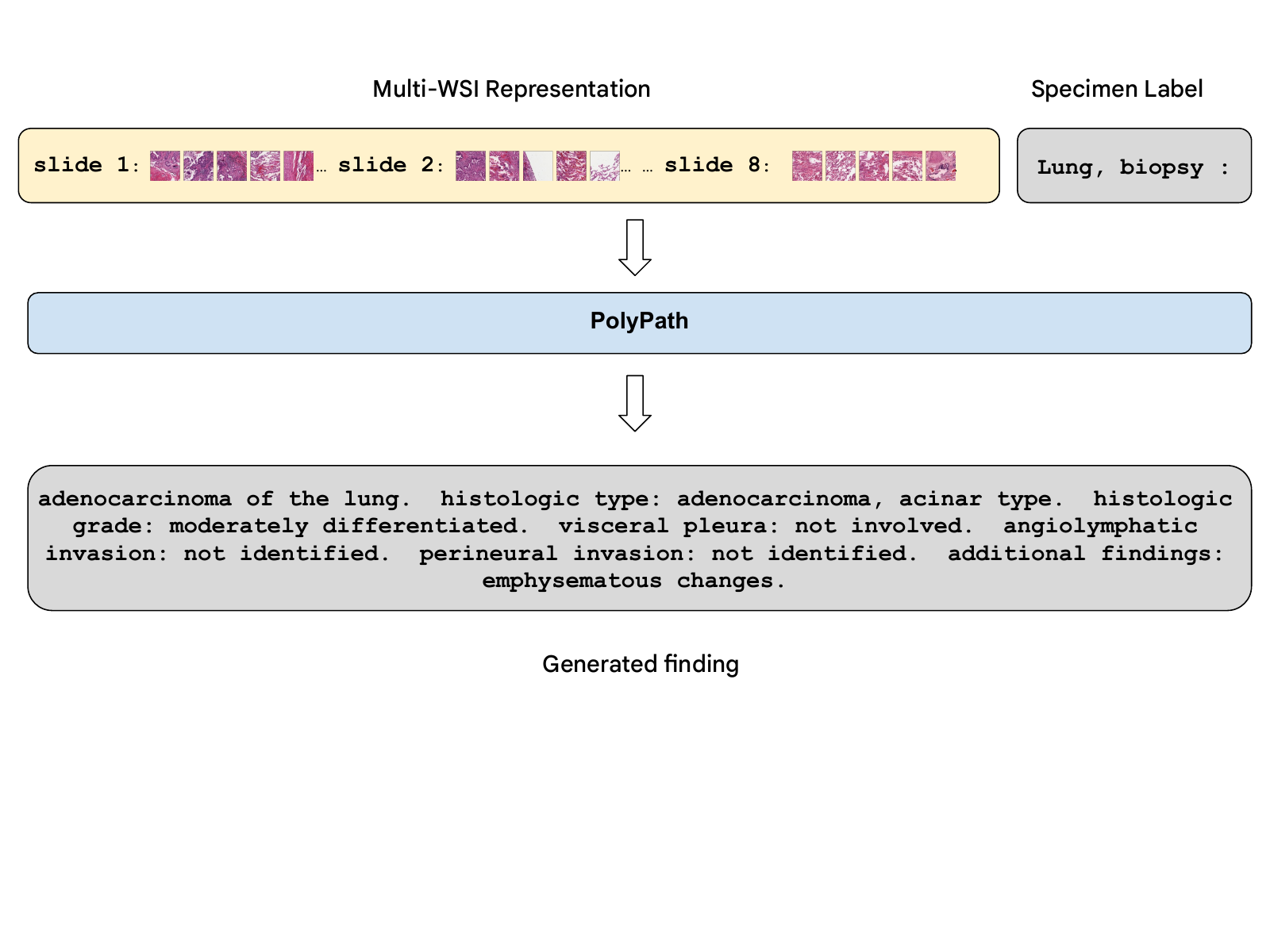}
    \caption{\small \textbf{Model overview.} \ourmodel is an LMM based on Gemini 1.5 Flash that takes in tissue-containing image patches from multiple slides for a part as well as the part-level specimen label, generating part-level report text across a wide range of tissue sites, procedures, and pathologies. The flow above reflects usage at inference time; during training, the model is trained to predict both the label and finding sections.}
    \label{fig:model}
\end{figure}

\section{Methods}
\subsection{Dataset}
\label{sec:dataset}
We build off of the recent data curation effort in~\citet{ahmed2024pathalign}, where the primary dataset consists of de-identified retrospective data of approximately 350K WSIs with associated pathology reports from a tertiary teaching hospital. The study was reviewed by Advarra IRB (Columbia, Maryland) and deemed exempt from further review as all data are retrospective and de-identified. The dataset is representative of the distribution of surgical and biopsy procedures, and findings read by pathologists in the U.S. in a large hospital system. The vast majority of slides are hematoxylin and eosin (H\&E) stained. This dataset was referred to as \textsc{DS1} in~\citet{ahmed2024pathalign}. \\ \\
As is typical for pathology reports, each case has an associated report text, corresponding to the final diagnosis (i.e. ``bottom-line'' text). These report texts are structured into part-level sections. In histopathology, a part refers to a specific tissue specimen submitted for pathology review. There may be multiple parts submitted for a given case, and for each part it is typical for multiple slides to be prepared for pathologist review. For each part-level section in the pathology report, there is a \emph{label} (description of anatomical site and surgical procedure) and a \emph{finding} (description of diagnostic findings), as illustrated in~\autoref{fig:model}. The \emph{label} can often include information that is non-inferable from the WSIs alone (e.g., related to specimen location or procedure). For each part, we associate its corresponding report text with the set of WSIs that are available for that part. Since the number of slides per part is a key variable in this study, we define the following dataset subsets consisting of parts with varying number of slides per part:
\begin{enumerate}
\setlength{\itemsep}{0pt}
\item \textsc{P1}: Parts with a single slide,
\item \textsc{P2-5}: Parts with 2 to 5 slides,
\item \textsc{P6-9}: Parts with 6 to 9 slides,
\item \textsc{P10+}: Parts with 10 or more slides.
\end{enumerate}
For this work, we resplit cases in the dataset into train, validation and test sets with the goal of achieving a similar distribution across part-categories within each split. For cases with a part in P1, we used the same train/validation/test splits as in~\citet{ahmed2024pathalign}. Parts in the remaining (multi-slide) part-categories were split into train/validation/test splits with ratios of approximately 70/20/10. The final part counts per split deviate slightly from the above ratio because we also ensure that each multi-part case has all of its parts in the same split.~\autoref{tab:dataset} summarizes the final numbers of parts in each split. We provide further details about distributions of tissues/sites for these categories in Supplemental~\autoref{fig:sitedistribution}.

\begin{table}[t]
\small
\centering
\caption{\textbf{Dataset Overview.} Number of parts per split for each part-category in the dataset.}
\begin{tabular}{@{}lcrrr@{}}
\toprule
 & & \multicolumn{3}{c}{Number of parts} \\
\cmidrule(lr){3-5}
Category & Slides per part & Train & Validation & Test \\
\midrule
\textsc{P1} & 1 & 36986 & 4561 & 4763 \\
\textsc{P2-5} & 2--5 & 49374 & 14541 & 7244 \\
\textsc{P6-9} & 6--9 & 4690 & 1381 & 594 \\
\textsc{P10+} & 10+ & 2479 & 825 & 405 \\
\bottomrule
\end{tabular}
\label{tab:dataset}
\end{table}

\subsection{Modeling}
\label{sec:modeling}
Our modeling approach consists of fine-tuning Gemini 1.5 Flash to generate part-level reports using all the WSIs for a given part as input. We utilized low-rank adaptation (LoRA)~\citep{hu2021lora}, a method for efficiently fine-tuning large pre-trained models by learning low-rank matrices that represent weight-changes while keeping the original model-weights frozen. LoRA was used with rank 16 and applied to all attention and feed-forward layers of the language model. The patch-level image encoder, a version specialized to medical modalities~\citep{yang2024advancing}, was pretrained and frozen. We refer to the entire multi-slide model -- frozen image encoder and tuned language model -- as \ourmodel. The input WSIs are represented as a sequence of non-overlapping, tissue-containing $768 \times 768$ image patches in row-major order for each slide (illustrated in~\autoref{fig:model}). We use all slides per part, sampling up to 50 slides without replacement for the approximately 0.5\% of parts where more than 50 slides were available. As in~\citet{ahmed2024pathalign}, we use patches from WSIs at 10X magnification. The median, 95th percentile, and maximum number of tissue-containing patches per part in our training data are 276, 2853, and $\sim$41K; see Supplemental~\autoref{fig:slidesandpatchesoverall} for the overall distribution of number of slides and tissue-containing patches per part. We perform all hyper-parameter tuning and checkpoint selection on the validation set, using the average of two NLG metrics -- ROUGE-L~\citep{lin2004rouge} and METEOR~\citep{banerjee2005meteor}. \\ \\
As a baseline, we also train a single-slide model (SS). For each slide from each part, we created a training example where the input is the set of patches from that single slide and the target is the overall part-level label and diagnostic finding. As in the \tthnoisy data split in~\citet{ahmed2024pathalign}, the single slide may not contain enough information to predict this part-level information. We consider two approaches for repurposing this single-slide model for inference across multiple slides: (1) \randommodel: we randomly pick a slide per part and generate a part-level report using SS. (2) \llmmodel: we generate report texts for all individual slides for a part with SS, and prompt a separate, general purpose LLM (Gemini Flash 1.5) to pick a single report text from the set that is consistent with the set as well as most clinically significant (Supplemental~\autoref{fig:prompt}). The source images are not shown to this LLM, only the generated report texts. To help the LLM pick good report texts, our prompt contains 50 examples for in-context learning. For each example, the input consists of report texts for all individual slides for a part generated with SS and the output consists of the report text from the input that has the closest NLG score (average of ROUGE-L and METEOR) to the ground-truth. Tuning for the prompt and selection of in-context examples was done using the validation set. We also tried a variant of \textsc{SS-LLM}, where the LLM is prompted to combine the single-slide report texts into a part-level report text, but found it inferior to random selection.

\subsection{Evaluation}
\label{sec:evaluation}
We perform automatic evaluations using common NLG metrics (ROUGE-L and METEOR) as well as human expert evaluations by board-certified pathologists. For both evaluations, we prompt the models with the label portion of the report text (indicating tissue type and procedure), and compare the findings output from the prompted models with the corresponding findings in the ground-truth text. \\ \\
For human-evaluations, two U.S. board-certified pathologists evaluated the findings generated by \ourmodel (multi-slide model), \randommodel (single-slide model), \llmmodel (single-slide model), as well as the original report text for a selected subset of 105 P2-5 examples. We did not perform extensive human-evaluation on parts from P6-9 and P10+ due to lower observed NLG metrics (see~\autoref{sec:nlg_evals}) for these part-categories and time needed for pathologist review of parts with many slides. We did perform human evaluation for a small subset ($n=16$) of P6-9, which confirmed that the lower NLG scores for this subset were indicative of the relatively lower quality according to expert review. \\ \\
For selecting a subset of P2-5 for human evaluation, tissue types were first categorized as common (colorectal, skin, cervix) or less common, and an equal number of common parts and uncommon parts were randomly sampled from the test split. For each part, pathologists were presented with the set of WSIs in a web-based digital pathology viewer along with the four candidate report texts. These texts were randomly ordered and pathologists were blinded to the source to avoid bias in interpretation. As in~\citet{ahmed2024pathalign}, report texts were rated individually on a five-point scale:
\begin{itemize}
\setlength{\itemsep}{0pt}
    \item 1 -- Completely inaccurate,
    \item 2 -- Partially accurate,
    \item 3 -- Mostly accurate but clinically significant error or omission,
    \item 4 -- Mostly accurate without clinically significant error or omission,
    \item 5 -- Highly accurate.
\end{itemize}
Additional details on the scoring instructions provided can be found in Supplemental~\autoref{tab:rubric}. These ratings were used both as an absolute assessment of the quality of each report text and to compare the relative quality of report texts generated by different models (\autoref{fig:pathologistresults}, Supplemental~\autoref{fig:preferencebaselines}). The mapping from absolute to relative ratings is shown in Supplemental~\autoref{tab:comparecategories}. We further stratified analysis by the categories \emph{normal}, \emph{mild}, and \emph{significant} for the test set examples. The \emph{mild} category includes findings involving a range of conditions such as \emph{inflammation}, \emph{adenomas}, and other non-cancerous conditions. The \emph{significant} category includes \emph{carcinomas}, \emph{dysplasia}, and other findings with immediate implications for clinical management.

\section{Results}
\subsection{NLG Results}
\label{sec:nlg_evals}
We computed ROUGE-L and METEOR scores for \ourmodel (multi-slide model), \randommodel and \llmmodel (single-slide models) on all part-categories using the finding from the original report text as ground truth. The results are summarized in~\autoref{tab:nlg_main}. The metrics indicate consistent improvements for \ourmodel over the single-slide model variants across all part-categories. We also observe that metrics are lower for part-categories with more slides per part, especially for P6-9 and P10+. This is likely due to a combination of such part-categories being more complex to diagnose as well as being relatively less well-represented in our training sets (see~\autoref{tab:dataset}). We also performed the above evaluation without prompting the models with the label, and using the generated label and finding as in~\citet{ahmed2024pathalign} for computing NLG metrics (Supplemental~\autoref{tab:nlgnoprompt}).

\begin{table}[t]
\small
\centering
\caption{\small \textbf{NLG results.} ROUGE-L and METEOR evaluations on the test-set for the captions generated by \randommodel, \llmmodel, and \ourmodel. In all cases, the models were prompted with the \emph{label} (which typically captures the tissue type and procedure), leading to an output consisting of the \emph{findings}. The ground truth is the corresponding part-level bottom-line diagnosis in the report.}
\begin{tabular}{@{} l *{3}{c} *{3}{c} @{}}
\toprule
 & \multicolumn{3}{c}{ROUGE-L} & \multicolumn{3}{c}{METEOR} \\
\cmidrule(lr){2-4} \cmidrule(lr){5-7}
Category & \textsc{SS-random} & \textsc{SS-LLM} & \ourmodel & \randommodel & \llmmodel & \ourmodel \\
\midrule
\textsc{P1} & 0.495 & 0.495 & \textbf{0.498} & 0.497 & 0.497 & \textbf{0.503} \\
\textsc{P2-5} & 0.448 & 0.452 & \textbf{0.488} & 0.450 & 0.454 & \textbf{0.500} \\
\textsc{P6-9} & 0.331 & 0.349 & \textbf{0.398} & 0.352 & 0.368 & \textbf{0.427} \\
\textsc{P10+} & 0.314 & 0.313 & \textbf{0.364} & 0.350 & 0.344 & \textbf{0.409} \\
\bottomrule
\end{tabular}
\label{tab:nlg_main}
\end{table}

\subsection{Pathologist Evaluation}
\label{sec:human_evals}
\begin{figure}[t]
\centering
  \begin{subfigure}[b]{\textwidth}
    \centering
    \includegraphics[scale=0.25]{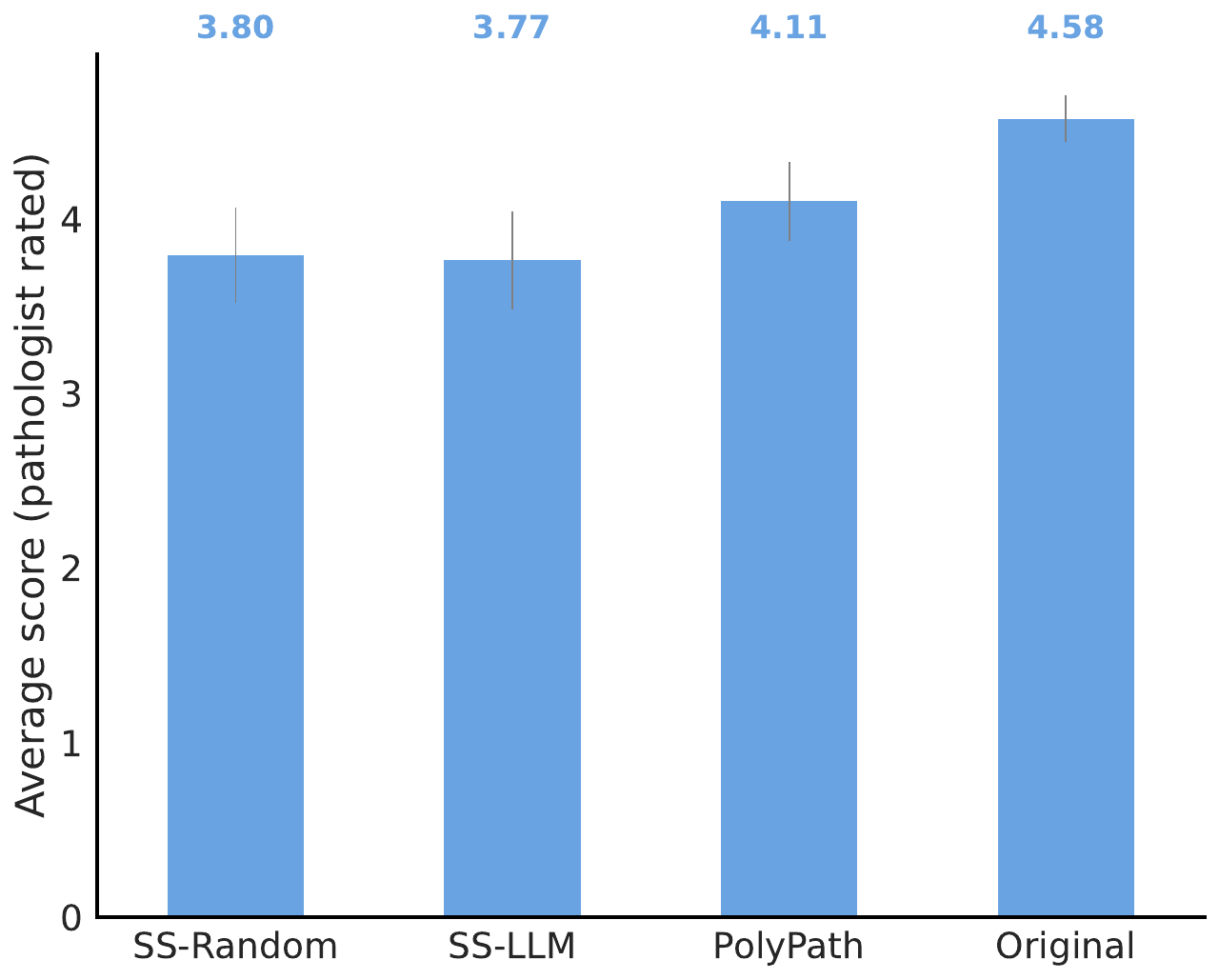}
    \caption{}
    \label{fig:avgscore}
  \end{subfigure}
  \\
  \begin{subfigure}[b]{\textwidth}
  \centering
  \includegraphics[scale=0.3]{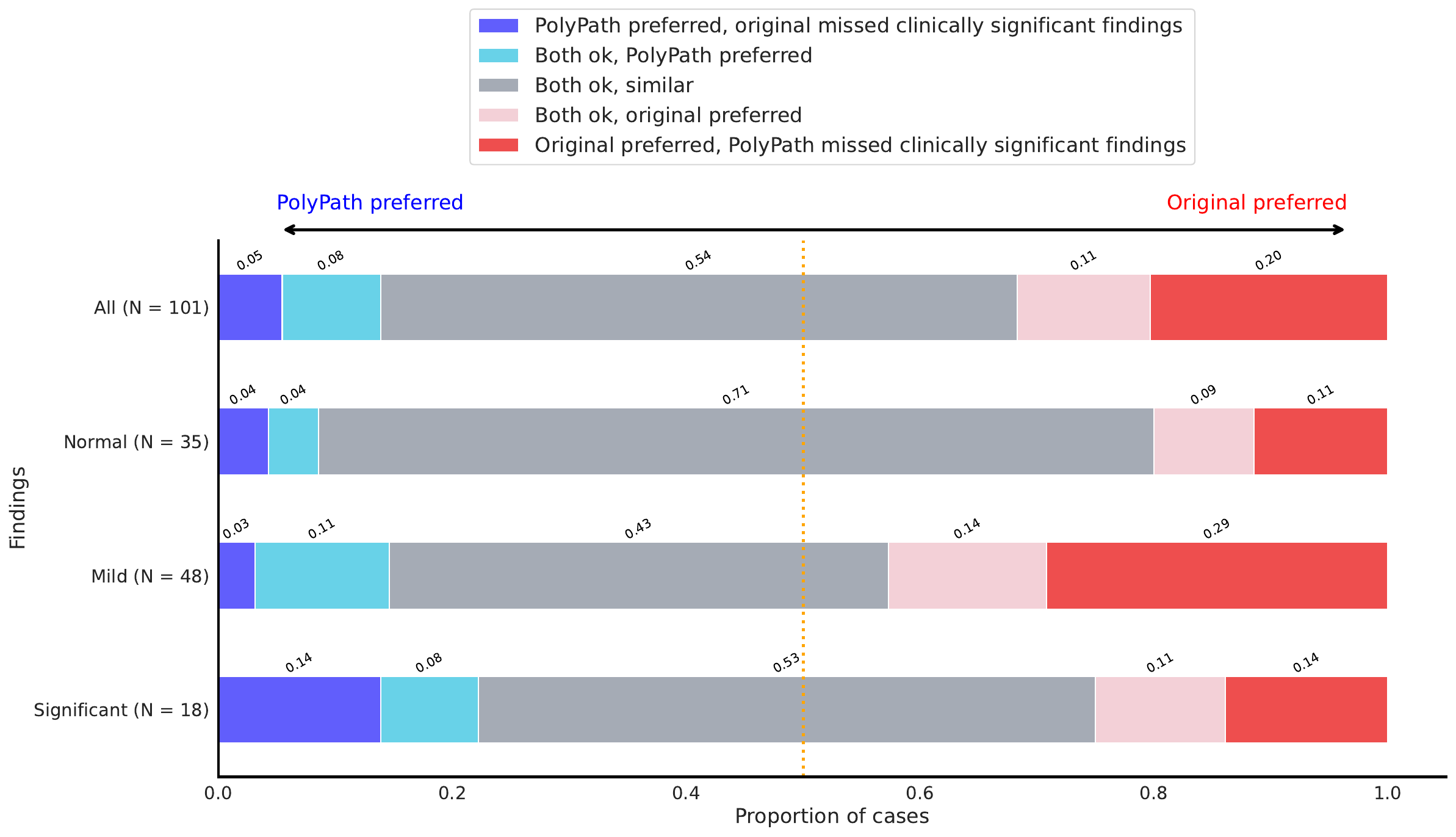}
  \caption{}
  \label{fig:preferenceoursvsoriginal}
  \end{subfigure}
\caption{\small \textbf{Pathologist evaluation.} Human evaluations conducted with two board-certified pathologists for 104 selected examples taken from the P2-5 subset of the test set. The scoring rubric is described in Supplemental~\autoref{tab:rubric}. (a) Average pathologist scores (higher is better) for report text generated by our three models (\randommodel, \llmmodel, \ourmodel) and the original bottom-line part-level diagnosis. Confidence intervals were generated via bootstrap resampling at the part-level with 10000 replicates. (b) Comparison of ratings for report text generated by \ourmodel vs the original report text for all parts as well as for subsets of parts with \emph{normal}, \emph{mild}, and \emph{significant} findings. For this plot, we exclude 3 examples for which at least one pathologist rated both the \ourmodel generated text and the original text at a score of 3 or lower.}
\label{fig:pathologistresults}
\end{figure}
We utilized the pathologist ratings for both individual and paired evaluations of models for the sample of 105 parts from P2-5. We dropped one case because one rater noted that more contextual information was required to reliably rate the associated texts, including the ground truth finding. For individual model evaluations, we computed the average score per model across parts ($n=104$) and raters ($n=2$). \ourmodel, with an average score of 4.11 out of 5 (95\% CI: [3.87, 4.33]), was rated significantly higher (Wilcoxon signed-rank test applied to differences in scores averaged across raters) than the single-slide model variants \randommodel ($p=0.003$) and \llmmodel ($p < 0.001$) with average scores of 3.80 (95\% CI: [3.53, 4.07]) and 3.77 (95\% CI: [3.49, 4.04]) respectively, while being rated significantly lower than the original report texts ($p < 0.001$), which averaged a score of 4.58 (95\% CI: [4.44, 4.71],~\autoref{fig:avgscore}). \\ \\
Comparing NLG metrics with pathologist ratings, we observe that both ROUGE-L and METEOR scores increased with increasing pathologist ratings (Supplemental~\autoref{fig:nlgcorr}). However, there was a large spread of NLG scores for any given pathologist rating, highlighting that, although NLG metrics can be conveniently calculated and are correlated with human expert evaluations, NLG metrics are not a substitute for human expert evaluations. \\ \\
For paired comparisons between models, pairs of individual ratings were mapped to categories as described in Supplemental~\autoref{tab:comparecategories}. In a comparison between \ourmodel and the original report text, output from \ourmodel was judged to be at least as good as the original report text for 68\% (95\% CI: [60\%, 76\%]) of examples (\autoref{fig:preferenceoursvsoriginal}). When we disaggregated this comparison by the severity of the original report text, i.e. \emph{normal}, \emph{mild}, \emph{significant} (\autoref{fig:preferenceoursvsoriginal}), \ourmodel-generated report texts were judged to be at least as good as the original report texts for 80\% of parts with normal findings, 57\% of parts with mild pathologic findings and 75\% of parts with significant pathologic findings (\autoref{fig:preferenceoursvsoriginal}). To characterize the inter-rater variability between the two raters, we include a version of this analysis disaggregated per pathologist reader in Supplemental~\autoref{fig:perpathologist}; the distribution of ratings per pathologist in Supplemental~\autoref{fig:ratingsperpathologist} and~\autoref{fig:stackedratings}; and inter-pathologist confusion matrices for scores for both the original report texts as well as generated report text from \ourmodel in Supplemental~\autoref{fig:disagreements}. We also conducted the paired comparison described above for \randommodel and \llmmodel (Supplemental~\autoref{fig:preferencerandomvsoriginal} and~\autoref{fig:preferencellmvsoriginal}), for which the generated report text was rated to be at least as good as the original report text at lower rates of 60\% (95\% CI: [52\%, 69\%]) and 61\% (95\% CI: [52\%, 70\%]) respectively.

\section{Discussion}
\label{sec:discussion}
In this work, we show that a large multimodal model with long-context capabilities can be used to make progress on the computationally challenging and clinically relevant task of generating report text descriptions from multiple WSIs. We show by means of expert pathologist review that the generated report text for parts with up to 2--5 slides are considered at least equivalent to the part-level report text in the original pathology reports 68\% of the time. \\ \\
In the majority of cases, pathologic interpretation requires review of multiple slides. This is especially true for specimens from complex pathologies such as cancer. In our dataset, which is representative of a typical clinical case distribution, at least 63\% of parts have more than one slide associated with them (Supplemental~\autoref{fig:sitedistribution}). While the current literature predominantly focuses on single-slide capabilities or multi-slide capabilities for a narrow range of anatomical sites, we believe general multi-slide modeling is a necessary step towards generalist diagnostic pathology models trained with supervision from pathology reports covering the wide range of tissue sites, procedures, and pathologies seen in clinical practice. Our experiments also suggest that the application of single-slide models naively may result in either lackluster performance or cascaded errors, further motivating the need for multi-slide modeling. From the clinical perspective, this may be intuitive: tasking multiple independent pathologists with reviewing and writing impressions for individual slides and then tasking a final pathologist with `combining' the impressions across these slides is suboptimal and fails to incorporate insights that can only be drawn by contextualizing findings across all slides in the case or part. \\ \\
In terms of evaluation, our data comparing absolute human expert ratings with NLG metrics also emphasize the importance of human ratings instead of relying exclusively on NLG comparisons with an original human-written diagnostic report (even when available). Specifically, we find that although NLG metrics were correlated with human ratings, with a monotonic increase in average NLG scores for both ROUGE-L and METEOR with higher human rating, the full spectrum of NLG scores across cases can span nearly the full range of 0--1 values. This likely stems from the fact that the NLG metrics are based on matching words or phrases (with some elementary semantic-matching), and there are often different ways of phrasing a report that corresponds to similar or identical clinical meaning. \\ \\
While our results for parts with up to 5 slides are promising, the NLG scores for parts with a higher number of slides are significantly lower. We believe this is in part due to relative scarcity of parts with more than 5 slides in our training data –- for example, while we have $\sim$49k parts with 2--5 slides for training, we only have 7.2k parts with more than 5 slides per part (\autoref{tab:dataset}). Another limitation is the lack of external validation; while our dataset is curated from a typical clinical setting, all of it comes from the same sites. As a result we have not assessed if the model performance generalizes to cases from different sites or geographical areas. \\ \\
While we explored the potential application of a long-context LMM for generating report text for parts with multiple slides, clinical reporting is typically done at a case-level, which can involve integration of information across multiple parts. Certain case-level tasks like cancer-staging are only possible when factoring in the findings for all the parts for a case. A trivial extension of our work to the case-level setting would be to run our part-level model for all the parts associated with a case and prompt an LLM to summarize part-level findings into case-level findings. However, similar to how in this work part-level models outperformed slide-level models at part-level interpretation, we hypothesize that case-level models (i.e. models that jointly process all slides in a case as inputs) could potentially enable more accurate case-level interpretation. We anticipate that patch-level, slide-level, and part-level pre-training will be important factors for improving the performance and data efficiency of case-level models. \\ \\
A further avenue of future work lies in exploring different ways of feeding image patches into the LMM that enable it to better utilize the spatial relationships between image-patches within and across slides. We also hope to probe the long-context abilities of LMMs such as Gemini even further. In our model, we made use of an aggressive mean-pooling strategy to reduce the number of visual tokens, by averaging all 256 tokens per patch. While we found this to be a useful strategy for efficient long-context modeling in our use case (pathology patches are typically more visually homogeneous at high-magnifications than natural images), it is possible that using more visual tokens would improve performance, at the cost of a longer context length and increased compute demands.

\section{Conclusion}
\label{sec:conclusion}
In this work we presented a system for generating part-level report text from parts with multiple WSIs. We expect that the combination of ongoing advances in patch-level pathology foundation models and long-context LMMs as well as large-scale digitization of pathology slide archives and reports will enable further development of clinically useful automated pathology interpretation systems.

\section*{Acknowledgements}
We thank Shawn Xu and Michael Howell for useful feedback on the manuscript. We thank the Google Research team for software and hardware infrastructure support as well as operations team members involved in the digitization and program management aspects related to this study; especially Melissa Moran, Robert MacDonald, Allen Chai, Robert Nagle, and Josh Pomorski. We also thank Kenneth Philbrick, Liron Yatziv, Can Kirmizi, and Rory Pilgrim for helpful technical discussions. We acknowledge James Wren and Colin Wageman for data-related discussions and thank the pathologists who reviewed model output for this study. We also thank Niels Olson and Arash Mohtashamian for contributions in initial aspects of establishing portions of the dataset used in this and prior studies.

\bibliographystyle{plainnat}
\bibliography{references}

\newpage
\appendix

\renewcommand\thefigure{\thesection.\arabic{figure}}    
\setcounter{figure}{0}    

\renewcommand\thetable{\thesection.\arabic{table}}
\setcounter{table}{0}

\section{Supplemental Tables}

\begin{table}[h]
\small
\centering
\caption{\small \textbf{NLG results for \emph{label+finding} generation, and \emph{label} prediction.} ROUGE-L and METEOR evaluations on all subsets of the test-set for the captions generated by \randommodel, \llmmodel, and \ourmodel. This is similar to the evaluation in Table 3, but without prompting the models with the \emph{label}, meaning that the models now output both the label as well as the findings (as done in~\citet{ahmed2024pathalign}). In (a), the ground truth is both the label and the finding, while in (b), we extract the \emph{label} portion of the texts and only compute NLG scores on the \emph{label} only. In (a), despite the need to predict more text, the numbers are significantly higher than in~\autoref{tab:nlg_main} because the models generally get the tissue type and procedure correct at a higher rate, as indicated by the numbers in (b). We note that \ourmodel performs better than the single-slide \textsc{SS} model (and the prior single-slide model in~\citet{ahmed2024pathalign}) on the single-slide subset as well when evaluated under the \emph{label+finding} style of evaluation in~\citet{ahmed2024pathalign}.}
\vspace{0.2cm}
\begin{tabular}{@{} l *{3}{c} *{3}{c} @{}}
\toprule
 & \multicolumn{3}{c}{ROUGE-L} & \multicolumn{3}{c}{METEOR} \\
\cmidrule(lr){2-4} \cmidrule(lr){5-7}
Category & \textsc{SS-random} & \textsc{SS-LLM} & \ourmodel & \randommodel & \llmmodel & \ourmodel \\
\midrule
P1    & 0.576 & 0.576 & \textbf{0.584} & 0.639 & 0.639 & \textbf{0.650} \\
P2-5  & 0.521 & 0.523 & \textbf{0.540} & 0.591 & 0.593 & \textbf{0.615} \\
P6-9  & 0.389 & 0.389 & \textbf{0.443} & 0.439 & 0.433 & \textbf{0.501} \\
P10+  & 0.331 & 0.344 & \textbf{0.400} & 0.383 & 0.388 & \textbf{0.458} \\
\Xhline{2.5\arrayrulewidth}
\end{tabular}
\vspace{0.1cm}

(a) NLG scores for generating \emph{label+finding}
\vspace{0.4cm}

\begin{tabular}{@{} l *{3}{c} *{3}{c} @{}}
\toprule
 & \multicolumn{3}{c}{ROUGE-L} & \multicolumn{3}{c}{METEOR} \\
\cmidrule(lr){2-4} \cmidrule(lr){5-7}
Category & \textsc{SS-random} & \textsc{SS-LLM} & \ourmodel & \randommodel & \llmmodel & \ourmodel \\
\midrule
P1    & \textbf{0.723} & \textbf{0.723} & 0.720 & 0.663 & 0.663 & \textbf{0.669} \\
P2-5  & 0.694 & 0.693 & \textbf{0.705} & 0.630 & 0.630 & \textbf{0.637} \\
P6-9  & 0.618 & 0.622 & \textbf{0.643} & 0.577 & 0.576 & \textbf{0.601} \\
P10+  & 0.536 & 0.561 & \textbf{0.562} & 0.504 & 0.517 & \textbf{0.524} \\
\Xhline{2.5\arrayrulewidth}
\end{tabular}
\vspace{0.1cm}

(b) NLG scores for \emph{label only}
\label{tab:nlgnoprompt}
\end{table}

\begin{table}[h]
\small
    \caption{\small \textbf{Comparison to PathAlign.} We compare NLG metrics for generated part-level report text for \pathalign~\citep{ahmed2024pathalign} and a version of \ourmodel trained and evaluated on the same data as \pathalign, i.e. trained on the single-slide \tthnoisy dataset in~\citet{ahmed2024pathalign} and evaluated on their corresponding single-slide test split.}
    \centering
    \begin{tabular}{lcc}
    \toprule
    Model & ROUGE-L & METEOR \\
    \midrule
    PathAlign & 0.579 & 0.612 \\
    PolyPath & 0.576 & 0.616 \\
    \bottomrule
    \end{tabular}
    \label{tab:pathaligncomparison}
\end{table}

\begin{table}[h]
\small
\caption{\small \textbf{Rating rubric.} Board-certified pathologists reviewed the set of slides per part and then rated model generated report text along with the original report texts on a scale of 1 to 5, with an option to indicate issues with quality or the need for more information. The set of texts per part were presented in random order and the raters were blinded to the source of the texts.}
\centering
\begin{tabular}{m{1.5cm} p{14cm}}
\toprule
\textbf{Score} & \textbf{Description and instructions} \\
\midrule
1 & Completely inaccurate
\begin{itemize}
\setlength{\itemsep}{0pt}
    \item may describe something that can occur in the specimen/tissue type pictured, but fundamentally incorrect, or may be the wrong tissue type or concept altogether.
\end{itemize} \\
\hline
2 & Partially accurate (i.e. related but wrong)
\begin{itemize}
\setlength{\itemsep}{0pt}
    \item The text might describe an entity that is related to images, and occurring in that specimen type, but images are definitively a different diagnostic entity.
    \item May accurately describe something that is seen on the images, but additional, essential info is missing or incorrect.
\end{itemize} \\
\hline
3 & Mostly accurate but with clinically \textbf{significant} error/omission
\begin{itemize}
\setlength{\itemsep}{0pt}
    \item The text is a very good match/description for the images, but something is incorrect or missing that may have clinical or diagnostic implications.
\end{itemize} \\
\hline
4 & Mostly accurate with clinically \textbf{insignificant} error/omission
\begin{itemize}
    \item The text is a very good match/description for the images, but there may be a minor, clinically \textbf{insignificant} aspect that is incorrect or missing
\end{itemize} \\
\hline
5 & Highly accurate
\begin{itemize}
\setlength{\itemsep}{0pt}
    \item For example, the diagnosis is accurate and acceptable, but doesn't capture all of the details
    \item The text is a great description of the images, with no obvious information missing or incorrect.
    \item Note that even a very short summary or a description of “no pathologic findings” can still belong in this score.
\end{itemize} \\
\hline
Need more info & This selection can be used if there is simply too little information in the text (or the images) to provide any reasonable quality assessment - please provide a very brief comment regarding the issue and/or what additional info you would need. \\
& *If the issue is that you can interpret the images to some extent, but need IHC or other studies to be more confident, please still provide a score based on your best interpretation of the available images with a “need more info comment” in the comments column. \\
\bottomrule
\end{tabular}
\label{tab:rubric}
\end{table}

\begin{table}[h]
\small
\centering
\caption{\small \textbf{Relative rating categories.} Categories for evaluation when comparing two texts as used in~\autoref{fig:preferenceoursvsoriginal} and Supplemental~\autoref{fig:preferencebaselines}.}
\begin{tabular}{lcc}
\toprule
\textbf{Category} & \textbf{Text 1 rating} & \textbf{Text 2 rating} \\
\midrule
Text 1 preferred & $\geq$4 & $\leq$3 \\
\hline
Both ok, text 1 preferred & 5 & 4 \\
\hline
\multirow{2}{*}{Both ok, same rating} & 4 & 4 \\
& 5 & 5 \\
\hline
Both ok, text 2 preferred & 4 & 5 \\
\hline
Text 2 preferred & $\leq$3 & $\geq$4 \\
\hline
Both with errors or omissions & $\leq$3 & $\leq$3 \\
\bottomrule
\end{tabular}
\label{tab:comparecategories}
\end{table}

\begin{table}[h]
\caption{\small \textbf{Qualitative examples.} Three examples of generated text with ratings and rationales from the expert pathologist raters for lower scores.}
\scriptsize
\centering
\begin{tabular}{l p{1cm} >{\centering\arraybackslash}p{1cm} p{2.5cm} p{2.5cm} p{2.5cm} p{2.5cm}}
\toprule
 & \textbf{Label} & \textbf{\#slides} & \textbf{Original finding} & \textbf{SS-random} & \textbf{SS-LLM} & \textbf{PolyPath} \\
\midrule
\textbf{Example 1} & kidney, nephrectomy & 6 & renal cell carcinoma histologic type: clear cell histologic grade (fuhrman): fuhrman grade 3/4 extent of tumor: limited to kidney adrenal gland: not submitted additional findings: chronic interstitial inflammation, benign renal cyst & renal cell carcinoma histologic type: clear cell. histologic grade (fuhrman): fuhrman grade ii. extent of tumor: limited to the kidney. & renal cell carcinoma histologic type: clear cell. histologic grade (fuhrman): fuhrman grade ii. extent of tumor: limited to the kidney. & renal cell carcinoma. histologic type: clear cell type. histologic grade (fuhrman): grade 3. extent of tumor: limited to kidney. adrenal gland: not submitted. additional findings: chronic inflammation. \\
\hline
Rating 1 & & & \multicolumn{1}{c}{5} & \multicolumn{1}{c}{4} & \multicolumn{1}{c}{4} & \multicolumn{1}{c}{5} \\ 
Rationale 1 & & & & \multicolumn{1}{c}{Grade 3} & \multicolumn{1}{c}{Grade 3} &  \\
\hline
Rating 2 & & & \multicolumn{1}{c}{5} & \multicolumn{1}{c}{4} & \multicolumn{1}{c}{4} & \multicolumn{1}{c}{5} \\
Rationale 2 & & & & \multicolumn{1}{c}{favor grade 3} & \multicolumn{1}{c}{favor grade 3} &  \\
\midrule
\textbf{Example 2} & breast mass, ultrasound-guided core biopsy & 3 & invasive ductal carcinoma. - histologic grade (nottingham histologic score) glandular/tubular differentiation: score = 2 nuclear pleomorphism: score = 2 mitotic rate: score = 1 overall grade: 5/9, grade 1. & invasive ductal carcinoma, histologic grade 2. tubule formation: 3/3. mitotic rate: 1/3. nuclear pleomorphism: 2/3. - largest contiguous focus. - venous / lymphatic invasion: not identified. - ductal carcinoma in situ (dcis): not identified. - lobular carcinoma in situ (lcis): not identified. & invasive ductal carcinoma, histologic grade 2. tubule formation: 3/3. mitotic rate: 1/3. nuclear pleomorphism: 2/3. - largest contiguous focus. - venous / lymphatic invasion: not identified. - ductal carcinoma in situ (dcis): not identified. - lobular carcinoma in situ (lcis): not identified. & invasive ductal carcinoma, nuclear grade 2. - largest contiguous focus. - venous / lymphatic invasion: not identified. - ductal carcinoma in situ (dcis): present, intermediate grade, solid and cribriform patterns. - lobular carcinoma in situ (lcis): not identified. \\
\hline
Rating 1 & & & \multicolumn{1}{c}{3} & \multicolumn{1}{c}{4} & \multicolumn{1}{c}{4} & \multicolumn{1}{c}{5} \\ 
Rationale 1 & & & \multicolumn{1}{c}{Grade 2 (3/2/1)} & \multicolumn{1}{c}{DCIS present}  & \multicolumn{1}{c}{DCIS present} &  \\
\hline
Rating 2 & & & \multicolumn{1}{c}{4} & \multicolumn{1}{c}{5} & \multicolumn{1}{c}{5} & \multicolumn{1}{c}{4} \\
Rationale 2 & & & Favor grade 2 with 3 points for tubule formation  & & & Would get myoepithelial stains, favor no DCIS \\
\midrule
\textbf{Example 3} & esophagus, biopsy &  3 & acute esophagitis. - no fugnal organisms identified by gms. & squamous mucosa with increased intraepithelial eosinophils (up to 20 per high power field). & squamous mucosa with increased intraepithelial eosinophils (up to 20 per high power field). & acute esophagitis with herpesvirus and fungal elements.  \\
\hline
Rating 1 & & & \multicolumn{1}{c}{5} & \multicolumn{1}{c}{2} & \multicolumn{1}{c}{2} & \multicolumn{1}{c}{3} \\ 
Rationale 1 & & & & Acute esophagitis with eosinophils & Acute esophagitis with eosinophils & No herpes or fungus \\
\hline
Rating 2 & & & \multicolumn{1}{c}{5} & \multicolumn{1}{c}{1} & \multicolumn{1}{c}{1} & \multicolumn{1}{c}{3} \\ 
Rationale 2 & & & & no increased intraepithelial eosinophils & no increased intraepithelial eosinophils & No herpes or fungus \\
\bottomrule
\end{tabular}
\label{tab:qualitativeexamples}
\end{table}

\clearpage
\section{Supplemental Figures}

\begin{figure}[h]
\centering
  \begin{subfigure}[b]{\textwidth}
    \centering
    \includegraphics[scale=0.35]{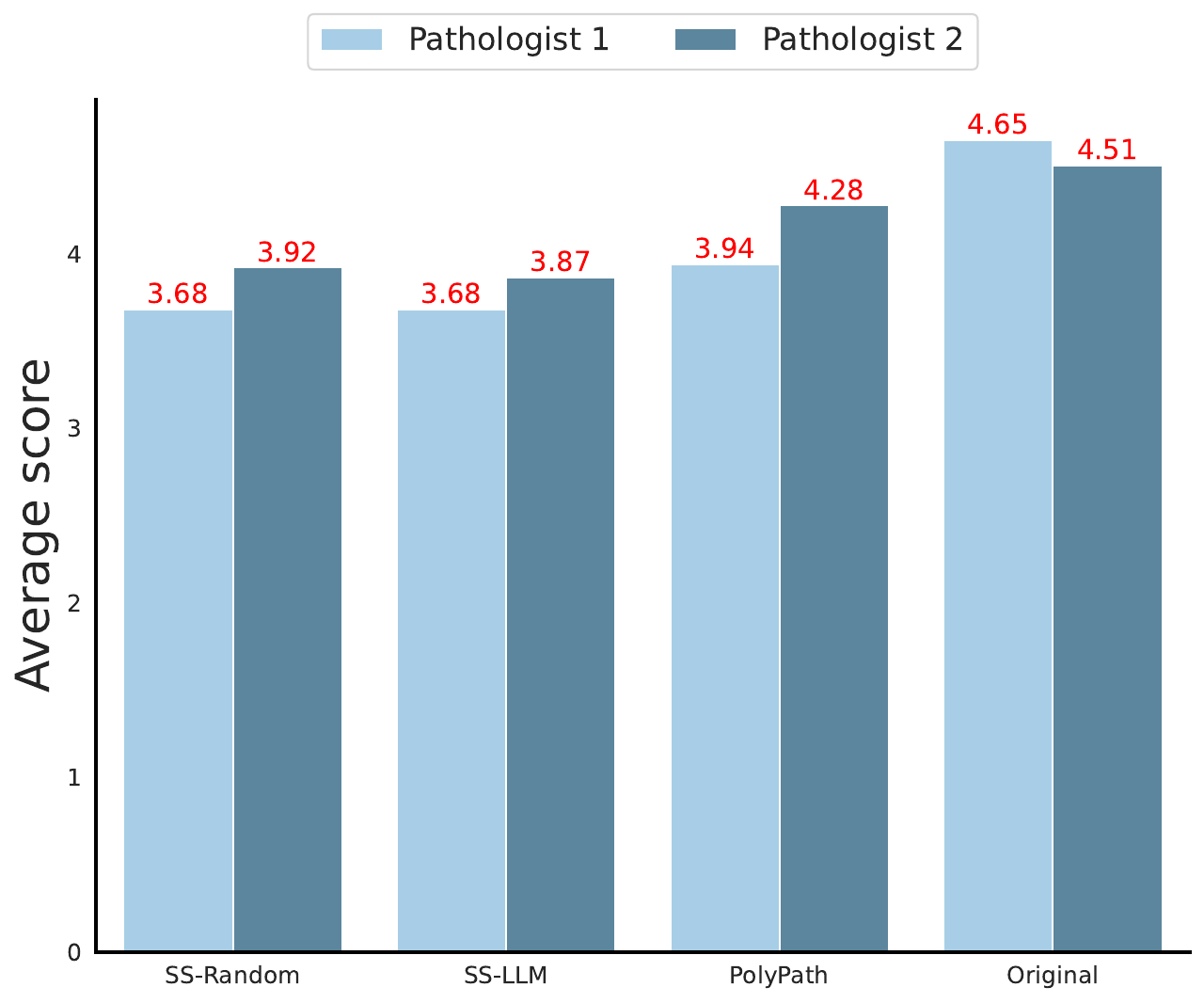}
    \caption{}
    \label{fig:averagescoreperpathologist}
  \end{subfigure}
  \\
  \begin{subfigure}[b]{\textwidth}
  \centering
  \includegraphics[scale=0.33]{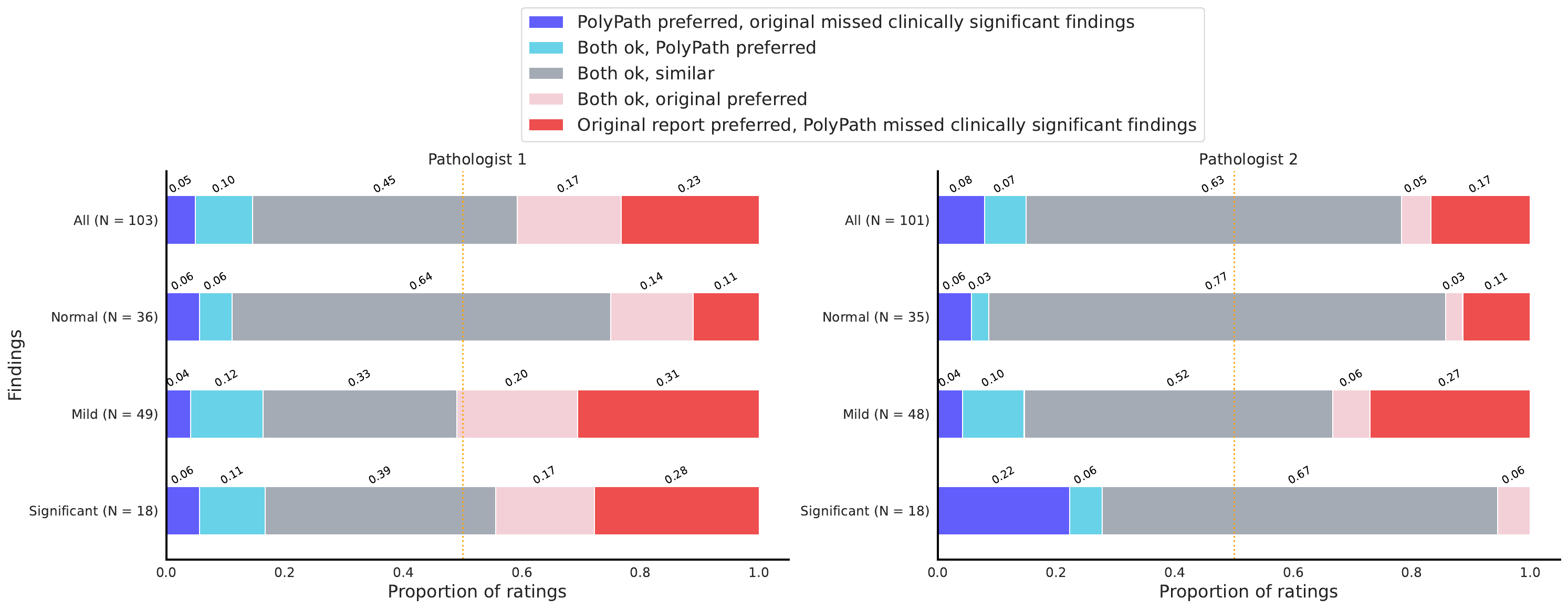}
  \caption{}
  \label{fig:preferenceperpathologist}
  \end{subfigure}
\caption{\small \textbf{Pathologist evaluations per individual rater, corresponding to~\autoref{fig:pathologistresults}.}  Plots for the P2-5 test split are shown for (a) average rating for generated report text from the \randommodel, \llmmodel and \ourmodel models as well as the original report text, and (b) preference plots for generated report text from \ourmodel vs. the original report texts.}
\label{fig:perpathologist}
\end{figure}

\begin{figure}[h]
\centering
  \begin{subfigure}[b]{\textwidth}
    \centering
    \includegraphics[scale=0.35]{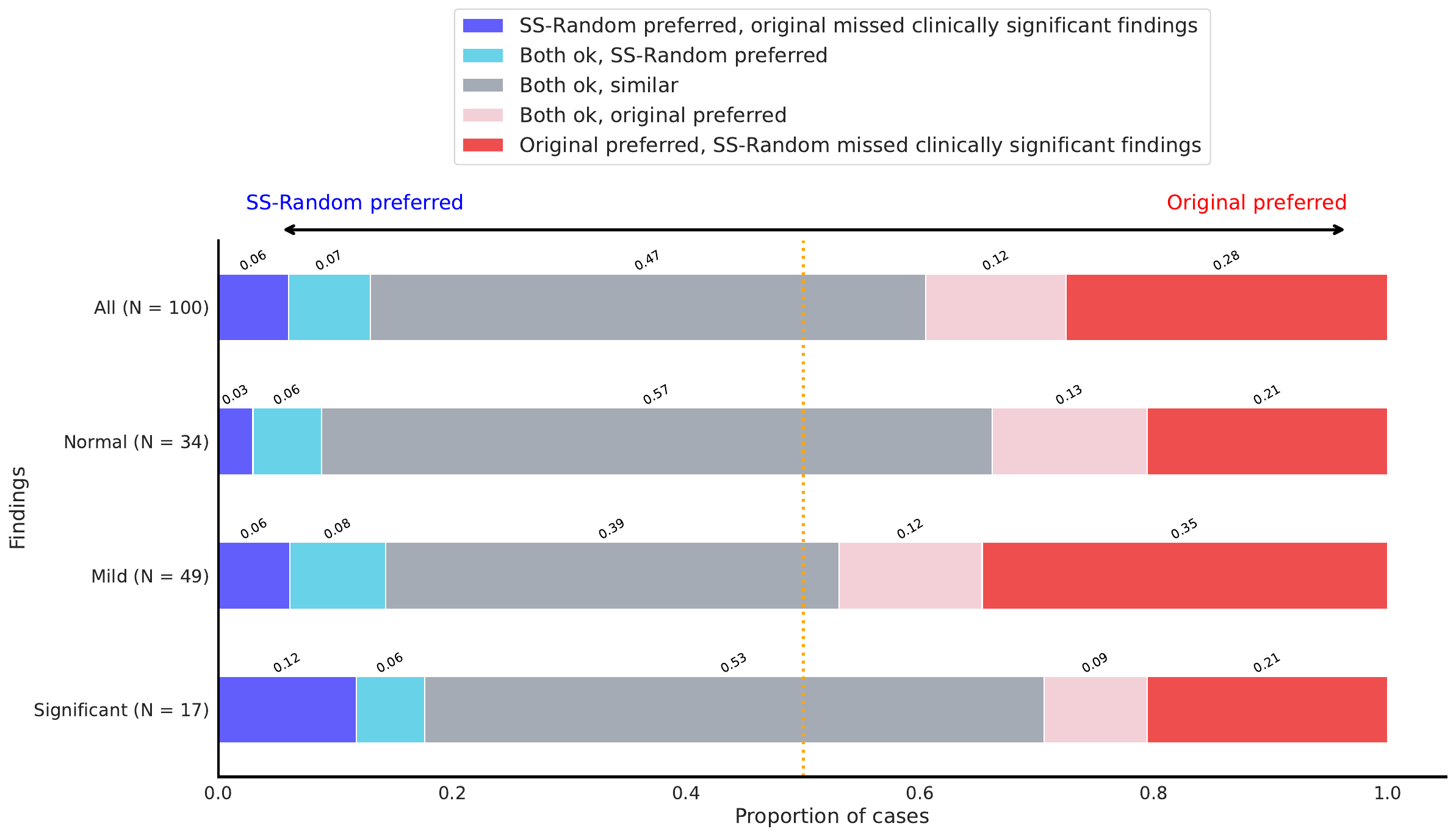}
    \caption{}
    \label{fig:preferencerandomvsoriginal}
  \end{subfigure}
  \\
  \begin{subfigure}[b]{\textwidth}
  \centering
  \includegraphics[scale=0.33]{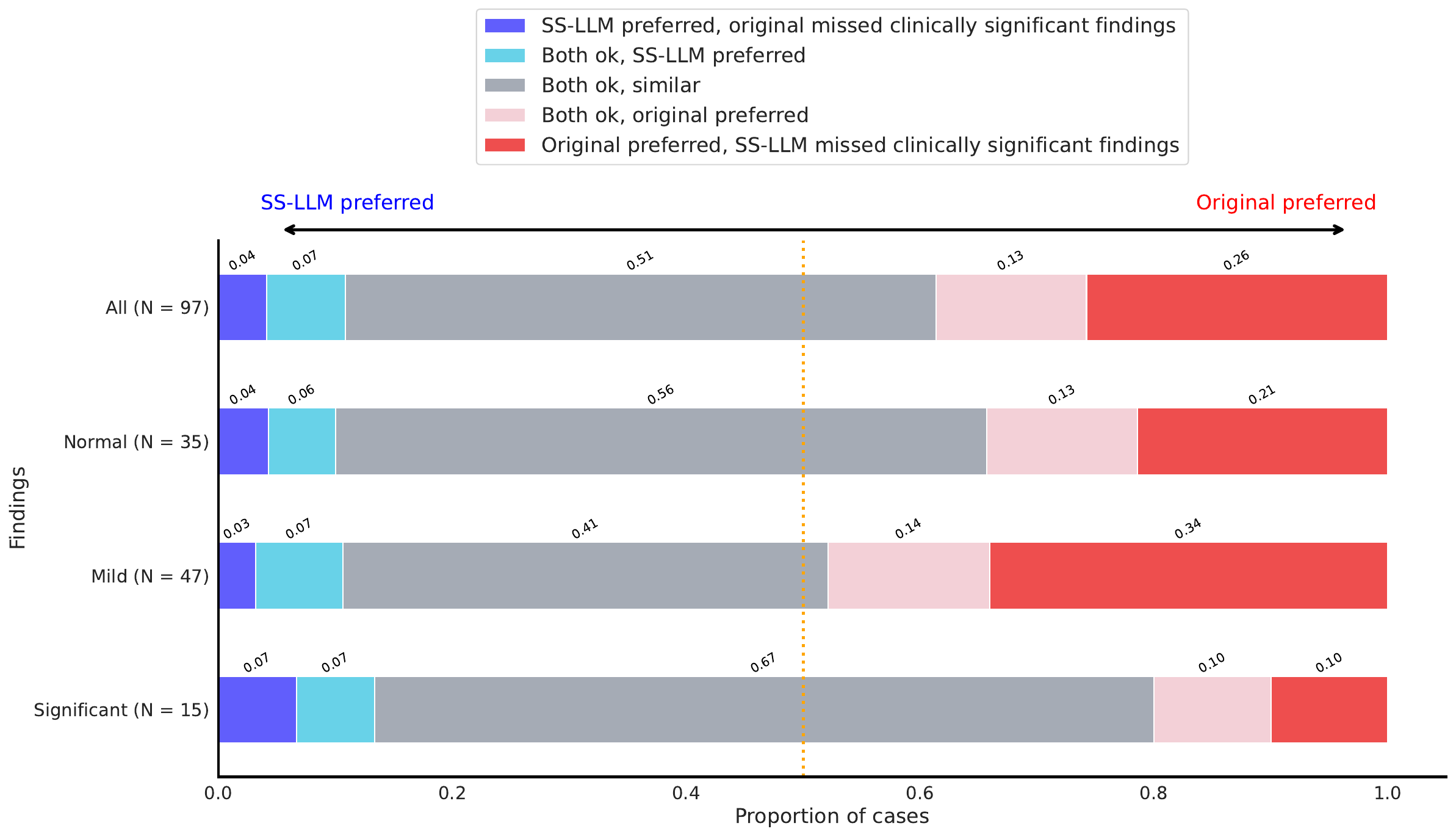}
  \caption{}
  \label{fig:preferencellmvsoriginal}
  \end{subfigure}
\caption{\small \textbf{Preference plots for single-slide models.} Comparison of ratings for report text generated by (a) \randommodel and (b) \llmmodel vs the original report text. Similar to the analysis in~\autoref{fig:pathologistresults}, we exclude examples where both the model output and the original text receive a score of 3 or lower by at least one pathologist.}
\label{fig:preferencebaselines}
\end{figure}

\begin{figure}
    \centering
    \includegraphics[scale=0.33]{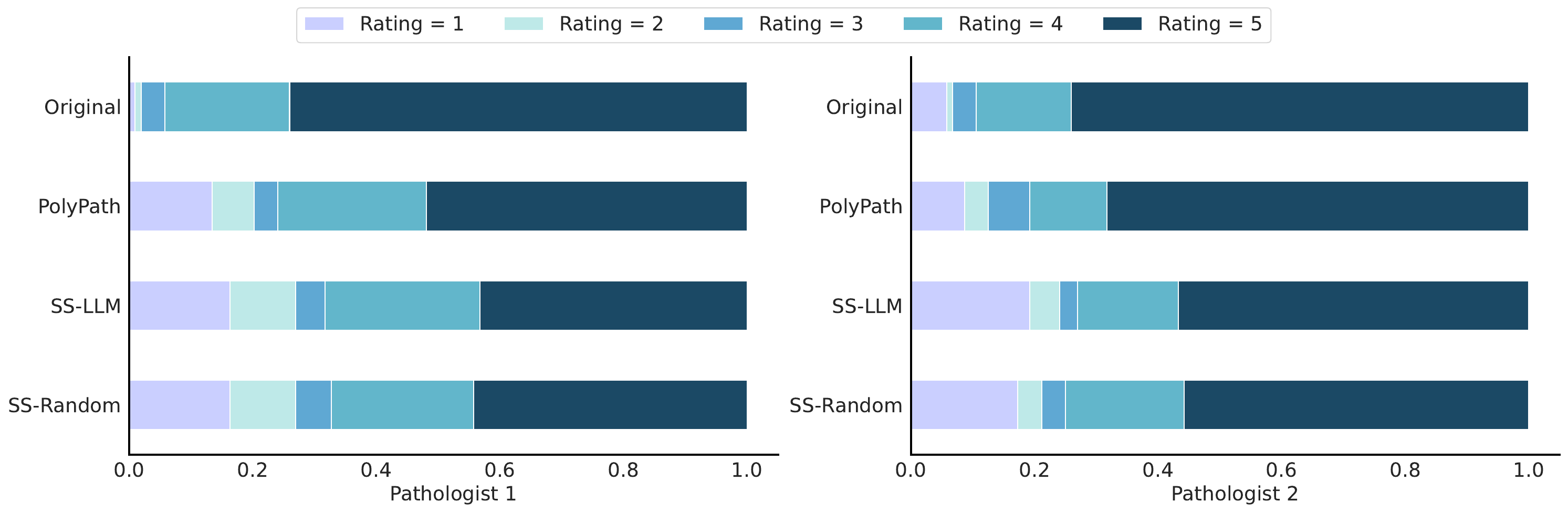}
    \caption{\small \textbf{Distribution of ratings.} Overall distribution of ratings per pathologist over the P2-5 test set.}
    \label{fig:ratingsperpathologist}
\end{figure}

\begin{figure}
    \centering
    \includegraphics[scale=0.33]{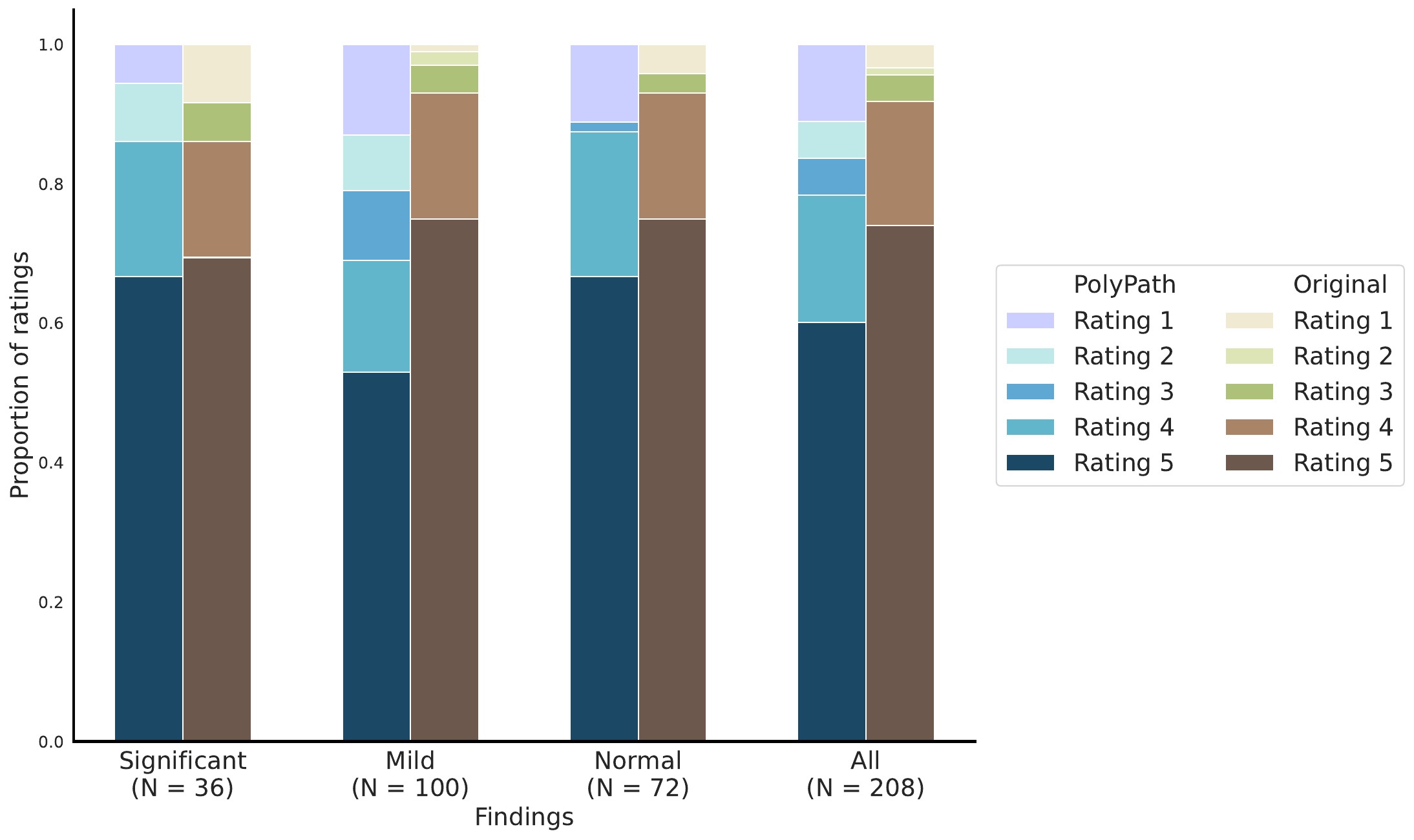}
    \caption{\small \textbf{Distribution of aggregated ratings for original report text vs. text generated from \ourmodel.} Aggregated ratings by category for generated text from \ourmodel and original report text. For parts across each finding category, the portion of ratings corresponding to each score is plotted for the \ourmodel generated text and original text, respectively.}
    \label{fig:stackedratings}
\end{figure}

\begin{figure}[h]
\centering
  \begin{subfigure}[b]{0.45\textwidth}
    \centering
    \includegraphics[scale=0.5]{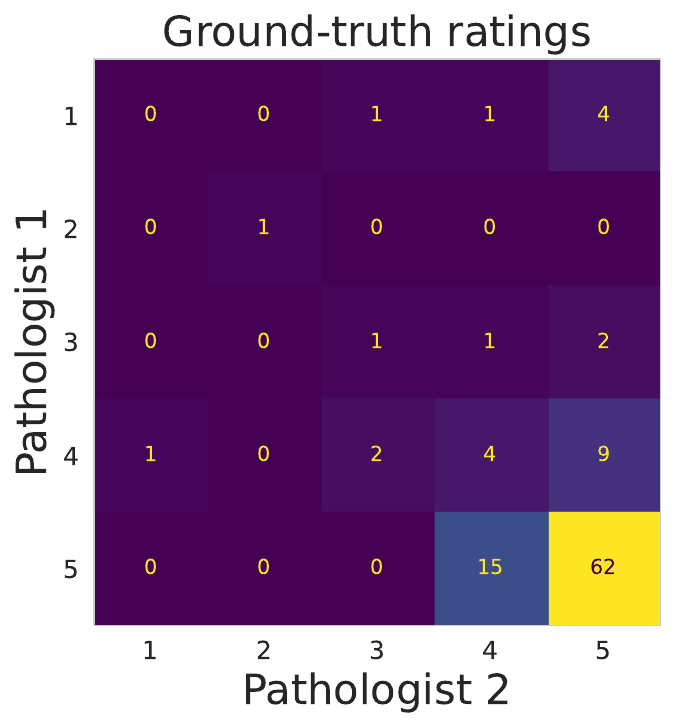}
    \caption{}
    \label{fig:disagreementsgt}
  \end{subfigure}
  \quad
  \begin{subfigure}[b]{0.45\textwidth}
  \centering
  \includegraphics[scale=0.5]{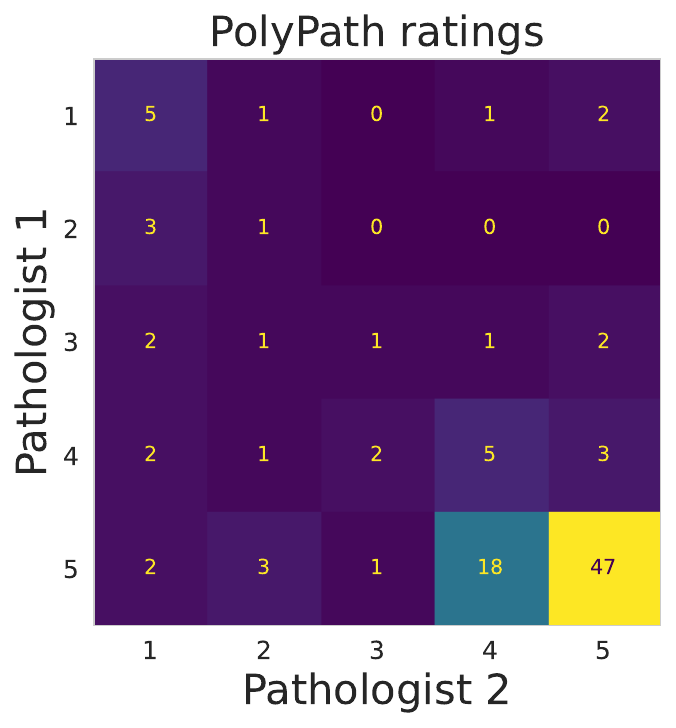}
  \caption{}
  \label{fig:disagreementsours}
  \end{subfigure}
\caption{\small \textbf{Disagreements between raters.} Confusion matrices of ratings between the two expert raters for the (a) ground truth report text, and (b) the text output from \ourmodel.}
\label{fig:disagreements}
\end{figure}

\begin{figure}[h]
\centering
\includegraphics[scale=0.55]{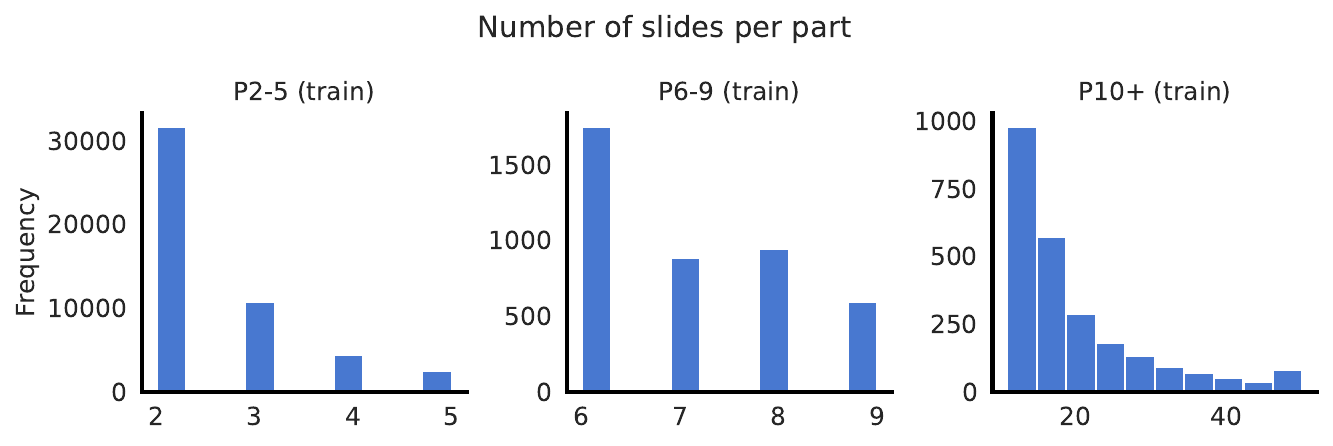} \\
\includegraphics[scale=0.55]{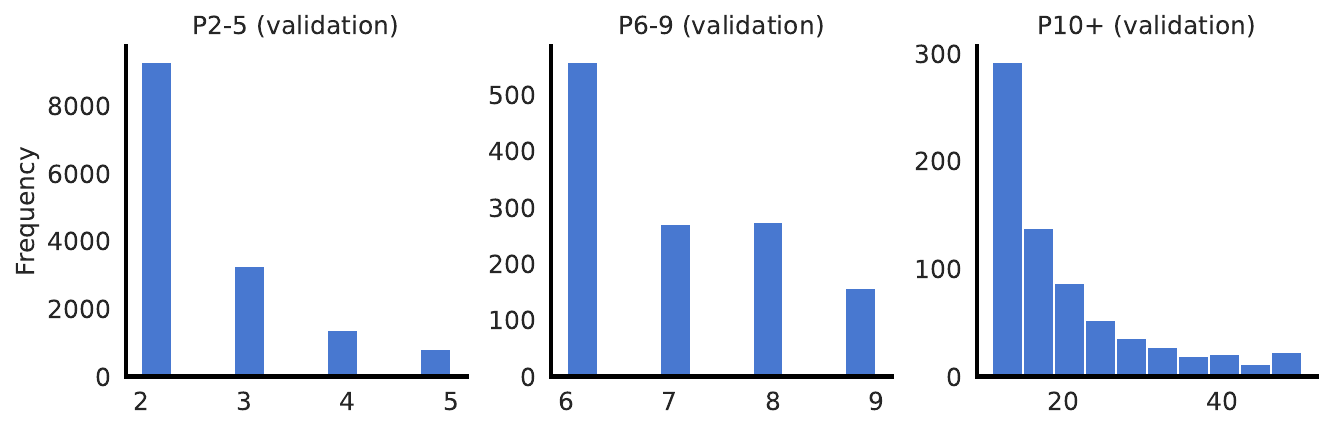} \\
\includegraphics[scale=0.55]{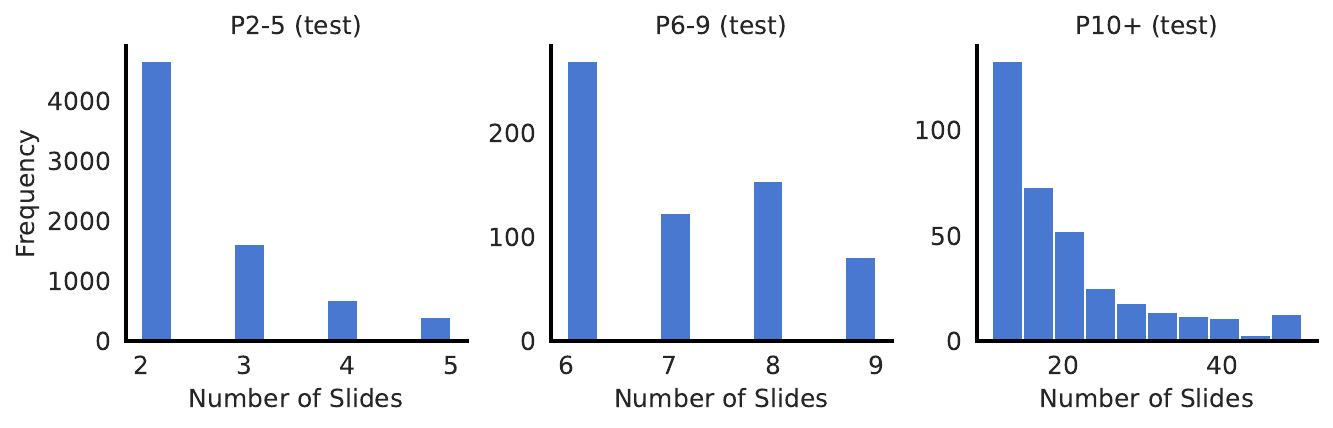}
\caption{\small \textbf{Distribution of number of slides per part across part-categories and splits.} Counts of number of slides per part for the multi-slide part-categories across (top) training, (middle) validation, and (bottom) test splits.}
\label{fig:slidedistribution}
\end{figure}

\begin{figure}[h]
\centering
  \begin{subfigure}[b]{\textwidth}
    \centering
    \includegraphics[scale=0.4]{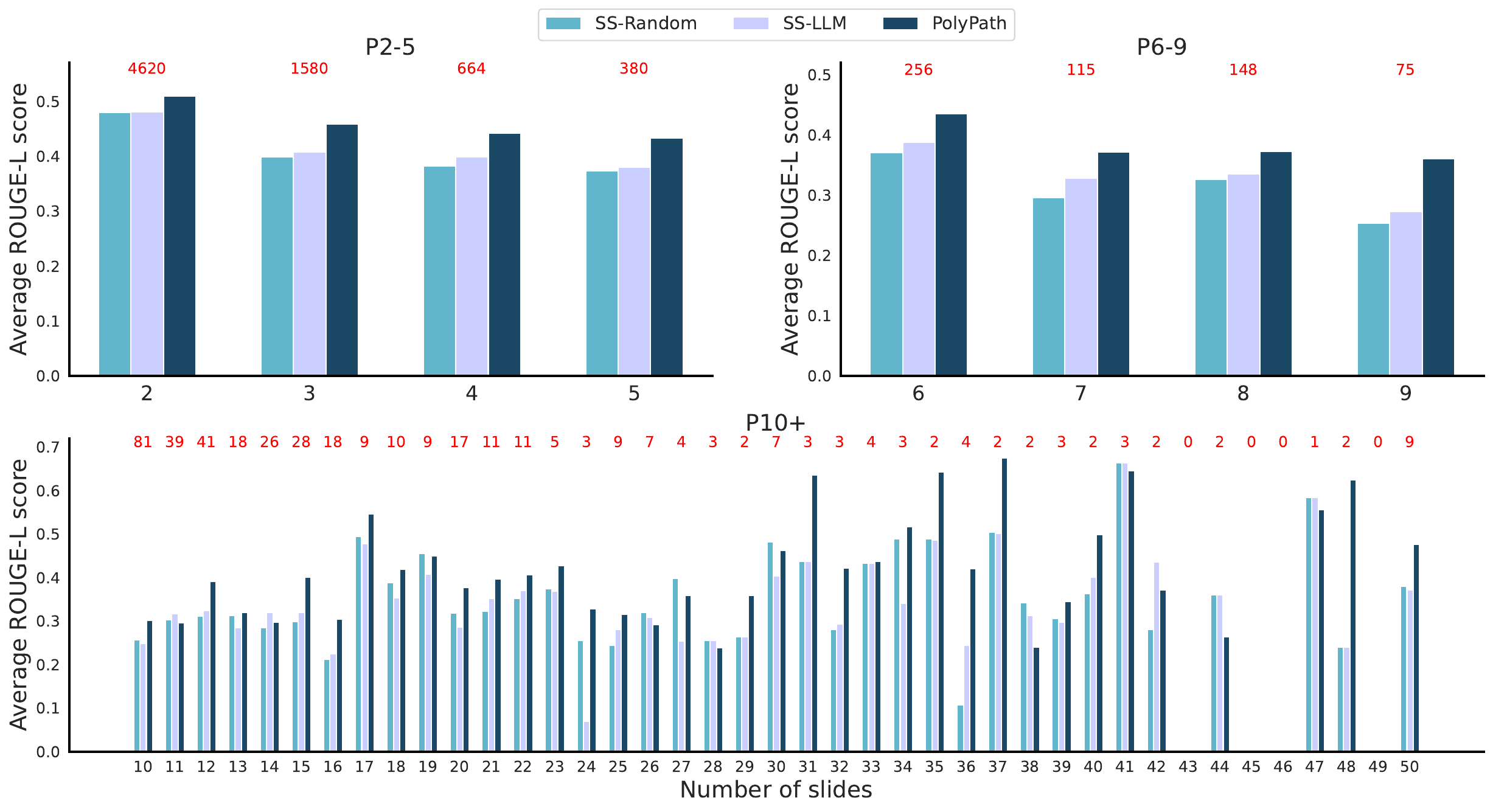}
    \caption{}
    \label{fig:rougepercategory}
  \end{subfigure}
  \\
  \begin{subfigure}[b]{\textwidth}
  \centering
  \includegraphics[scale=0.4]{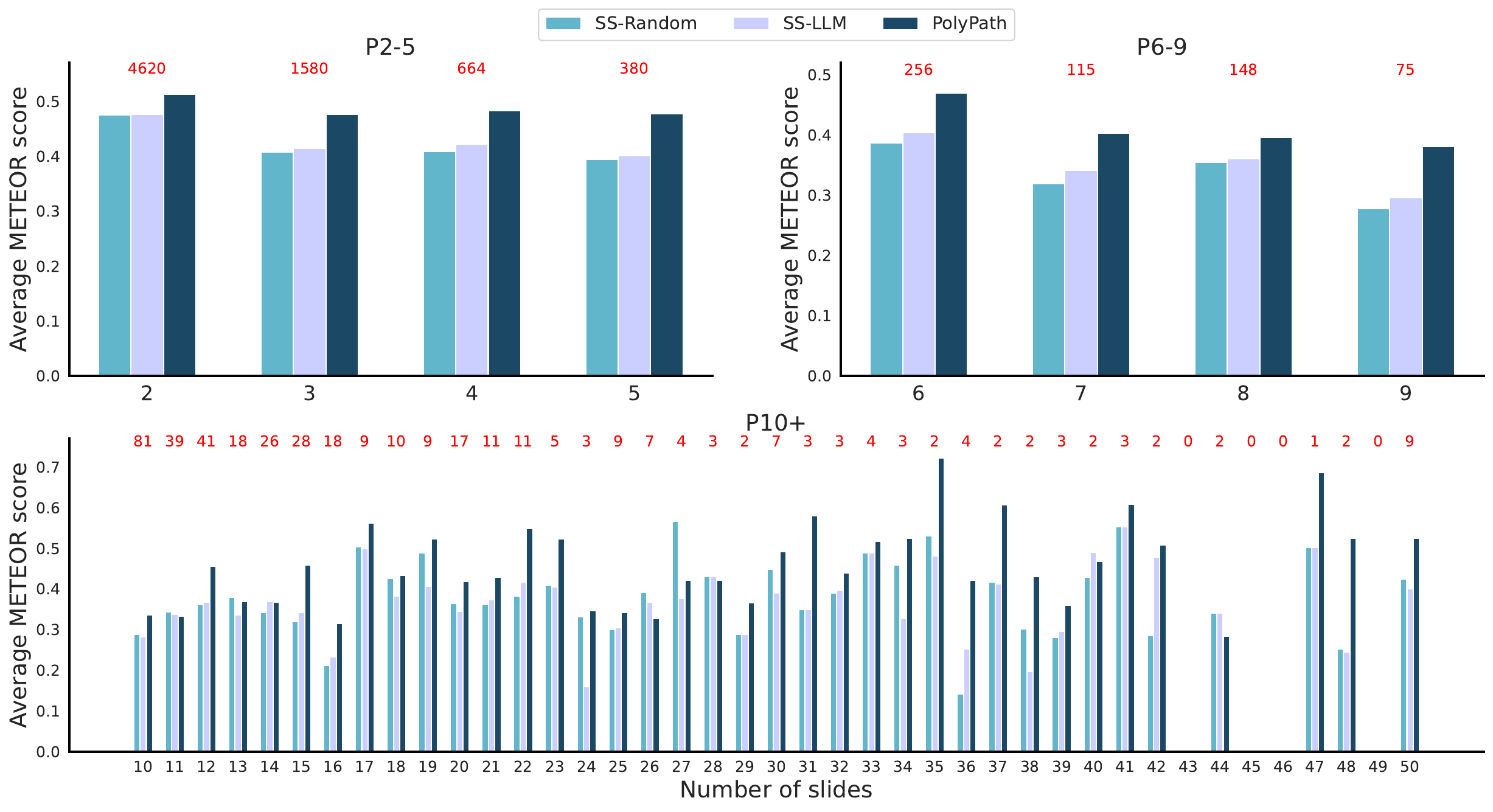}
  \caption{}
  \label{fig:meteorpercategory}
  \end{subfigure}
\caption{\small \textbf{NLG scores per number of slides in a part across part-categories in the test set.} Average NLG scores (a) ROUGE-L and (b) METEOR, for all models disaggregated by the number of slides in a part. The red numbers above each group corresponds to the number of examples corresponding to the number of slides per part.}
\label{fig:nlgpercategory}
\end{figure}

\begin{figure}[h]
\centering
\includegraphics[scale=0.4]{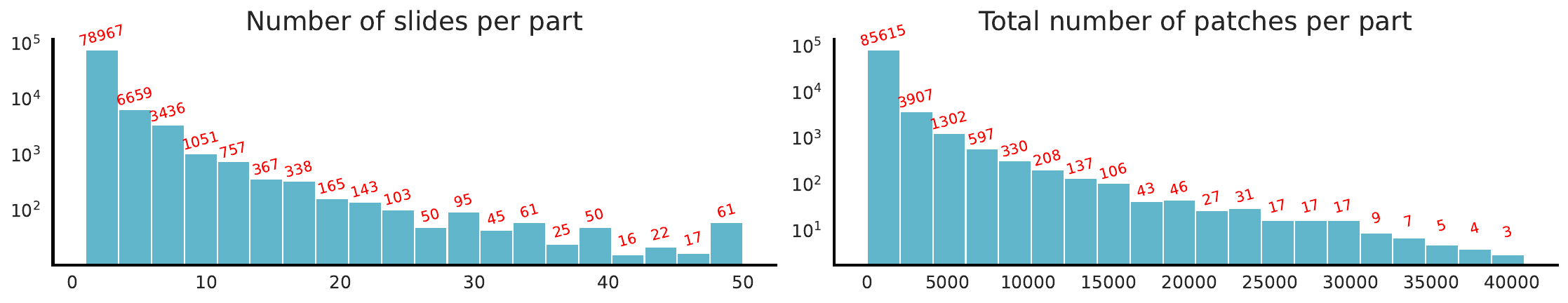} \\
\caption{\small \textbf{Overall distribution of number of slides and total number of patches per part.} We plot the distribution of (left) the number of slides per part in our training data with log-scale counts; (right) number of tissue-patches input to the PolyPath model per part. For the small ($\sim$0.5\%) of parts where the number of slides exceeds 50, we sample 50 slides without replacement.}
\label{fig:slidesandpatchesoverall}
\end{figure}

\begin{figure}[h]
\centering
\includegraphics[scale=0.4]{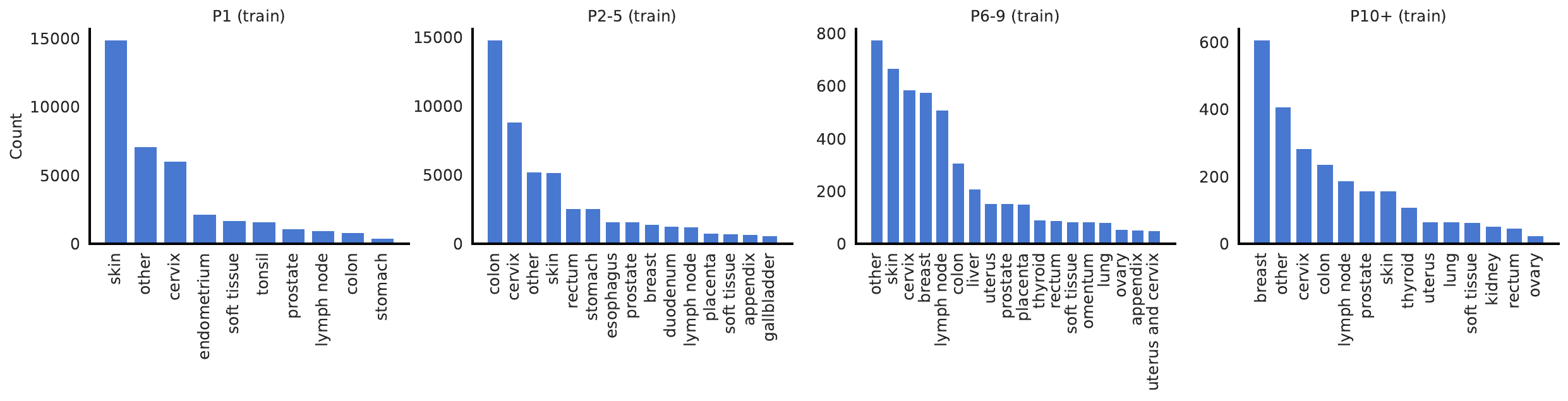} \\
\includegraphics[scale=0.4]{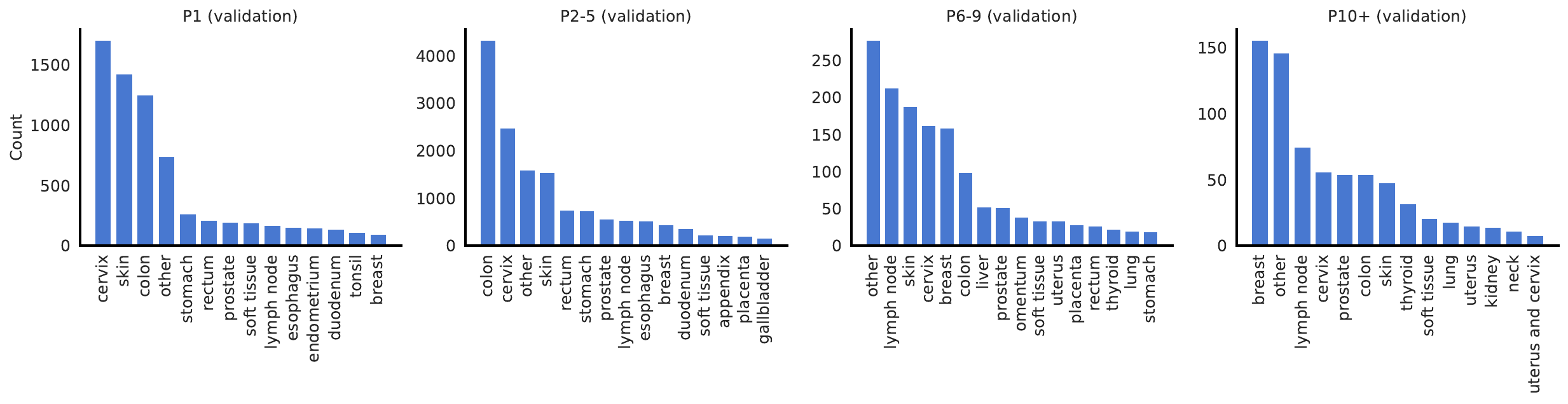} \\
\includegraphics[scale=0.4]{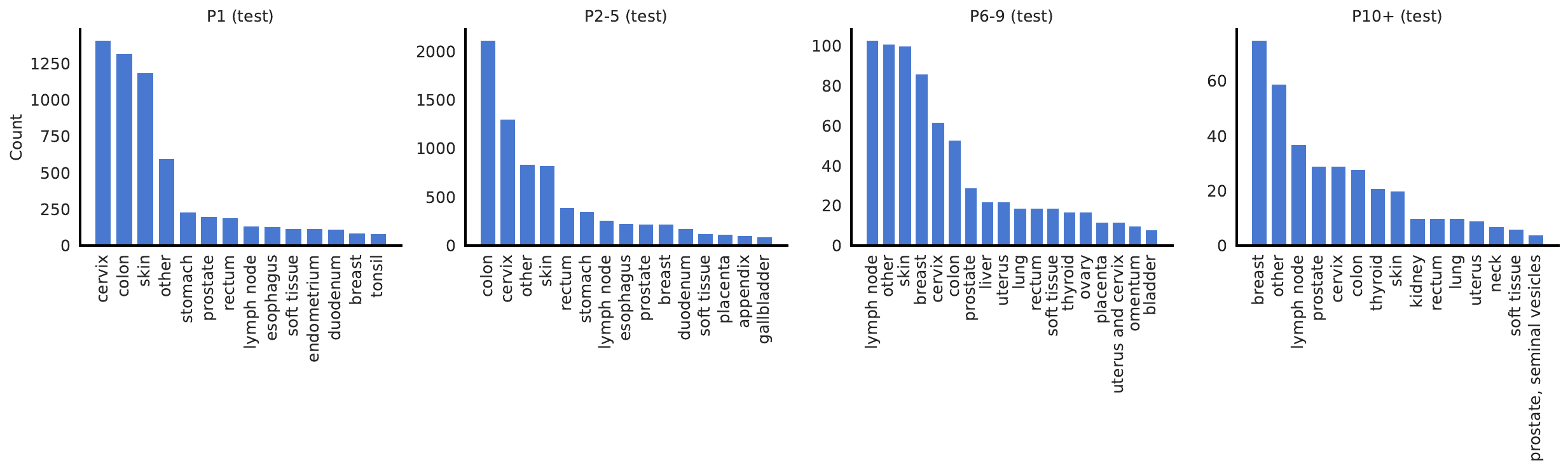}
\caption{\small \textbf{Distribution of tissue-counts over splits.} Distributions of tissue-counts over train/validation/test splits across the different part-categories. We used Gemini 1.5 Flash to identify a tissue/organ/site name from the labels corresponding to the parts, with spot-checking to test for quality. We then grouped all parts with a frequency of less than or equal to 1\% of the total count per (split, part-category) into “other”. These consist of less common tissues in our data, such as tooth, mandible, arteries, stapes, etc.}
\label{fig:sitedistribution}
\end{figure}

\begin{figure}[h]
\centering
\includegraphics[scale=0.5]{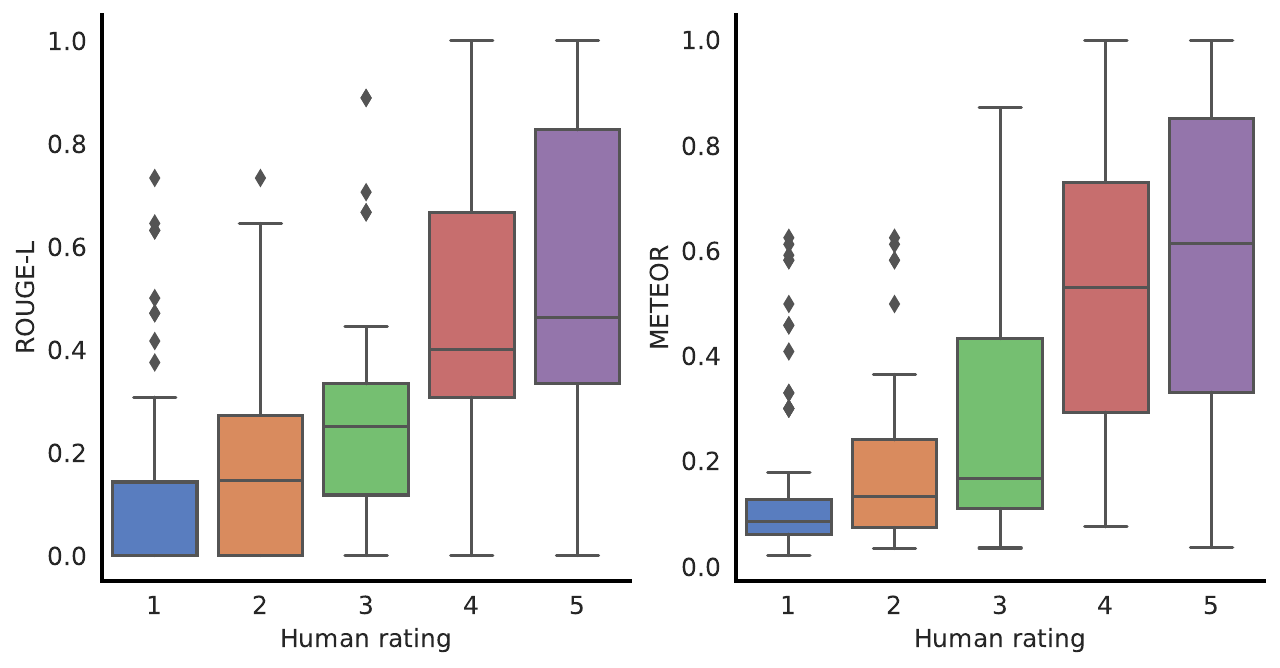}
\caption{\small \textbf{NLG scores and expert pathologist ratings.} We inspect the relationship between the human expert ratings and the NLG scores by plotting boxplots per human rating for (left) ROUGE-L and (b) METEOR scores. We aggregate the set of ratings for \randommodel, \llmmodel, and \ourmodel. Since the NLG metrics are relative to the ground truth and the human ratings are absolute, we drop reads where the original text was rated below 4, corresponding to $n = 191$ (down from a total number of 208 initial reads).}
\label{fig:nlgcorr}
\end{figure}

\begin{figure}[h]
\scriptsize
\begin{verbatim}
You are given a set of diagnosis notes for a set of pathology slides from a case.
Output a version that is most consistent and clinically significant.
If there is only one note, output that one.
Follow the examples below. Be sure to end with the <END> tag as in the examples.

*Example 1*
*Note 1* superficial basal cell carcinoma.
*Note 2* superficial basal cell carcinoma.
*Note 3* basal cell carcinoma, superficial type.
*Response*
superficial basal cell carcinoma.
<END>

*Example 2*
*Note 1* high-grade squamous intraepithelial lesion (cin ii).
*Note 2* benign endocervical tissue with squamous metaplasia; no dysplasia identified.
*Response*
high-grade squamous intraepithelial lesion (cin ii).
<END>

...
*Example 36*
*Note 1* gastric antral and fundic type mucosa with mild chronic gastritis :
    stomach, biopsy x2 (b-d): gastric antral and fundic type mucosa with 
    mild chronic gastritis.
*Note 2* chronic gastritis with intestinal metaplasia; negative for dysplasia.
 - helicobacter pylori-like organisms are not identified by routine and special stains.
*Note 3* gastric antral and fundic type mucosa with chronic inactive gastritis.
 - a warthin-starry stain for helicobacter pylori is negative.
*Response*
chronic gastritis with intestinal metaplasia; negative for dysplasia.
 - helicobacter pylori-like organisms are not identified by routine and special stains.
<END>

...
*Example 49*
*Note 1* gastric and squamous mucosa with no diagnostic alteration; 
    negative for intestinal metaplasia and dysplasia.
*Note 2* squamocolumnar junction mucosa with mild chronic inflammation; 
    negative for intestinal metaplasia and dysplasia.
*Note 3* squamocolumnar junction mucosa with mild chronic inflammation; 
    negative for intestinal metaplasia and dysplasia.
 - gastric mucosa with mild chronic gastritis.
*Note 4* squamocolumnar junction mucosa with mild chronic inflammation; 
    negative for intestinal metaplasia and dysplasia.
 - gastric mucosa with mild chronic gastritis.
*Note 5* squamocolumnar junction mucosa with mild chronic inflammation; 
    negative for intestinal metaplasia and dysplasia.
 - gastric cardia-type mucosa with chronic inactive gastritis.
*Response*
gastric and squamous mucosa with no diagnostic alteration; 
    negative for intestinal metaplasia and dysplasia.
<END>

*Example 50*
*Note 1* one lymph node, negative for metastatic carcinoma (0/1).
*Note 2* one lymph node, negative for metastatic carcinoma (0/1).
*Note 3* one lymph node, negative for metastatic carcinoma (0/1).
*Response*
one lymph node, negative for metastatic carcinoma (0/1).
<END>
\end{verbatim}
\caption{\small \textbf{Prompt used for \llmmodel.} The full prompt includes 50 examples for in-context learning, only a subset of which are shown here for space. For each example, the input consists of report texts for all individual slides per part generated with the SS model (\emph{Note 1, Note 2, …}) and the output (\emph{Response}) consists of the report text from the input that has the closest NLG score (average of ROUGE-L and METEOR) to the ground-truth. Note that although the prompt is not explicit about selecting one of the input texts, the 50-few shot examples provided in the prompt all involve picking one of the existing report texts rather than generating a new synthetic one.}
\label{fig:prompt}
\end{figure}

\end{document}